\def\year{2022}\relax

\documentclass[letterpaper]{article} 
\usepackage{aaai21}  
\usepackage{times}  
\usepackage{helvet} 
\usepackage{courier}  
\usepackage[hyphens]{url}  
\usepackage{graphicx} 
\usepackage{natbib}  
\usepackage{caption} 
\usepackage{color, colortbl}
\usepackage{subfig}
\newcommand{\vs}{\textit{vs}. }
\newcommand{\ie}{\textit{i}.\textit{e}.}
\newcommand{\eg}{\textit{e}.\textit{g}.}

\newcommand{\aka}{\textit{a}.\textit{k}.\textit{a}.}
\usepackage{xcolor}
\usepackage{kotex}
\usepackage{blindtext}

\usepackage{gensymb}
\usepackage{amsmath}
\usepackage{amssymb}
\usepackage{amsthm}
\usepackage{exscale}
\usepackage{textcomp}		
\usepackage{enumitem}

\usepackage{array}
\usepackage{tabulary}
\usepackage{multirow}
\usepackage{ctable} 
\usepackage{booktabs}	
\usepackage[toc,title,page]{appendix}

\usepackage[switch]{lineno}

\usepackage[flushleft]{threeparttable}
\usepackage[normalem]{ulem}

\title{Video as Conditional Graph Hierarchy for Multi-Granular Question Answering}
\author {
    Junbin Xiao, 
    Angela Yao, 
    Zhiyuan Liu, 
    Yicong Li, 
    Wei Ji, 
    Tat-Seng Chua \\
}
\affiliations {
    Department of Computer Science, National University of Singapore\\
    \{junbin, ayao, chuats\}@comp.nus.edu.sg,
    \ acharkq@gmail.com,
    \ liyicong@u.nus.edu,
    \ jiwei@nus.edu.sg
}

\begin{document}

\maketitle

\begin{abstract}
Video question answering requires the models to understand and reason about both the complex video and language data to correctly derive the answers. Existing efforts have been focused on designing sophisticated cross-modal interactions to fuse the information from two modalities, while encoding the video and question holistically as frame and word sequences. Despite their success, these methods are essentially revolving around the sequential nature of video- and question-contents, providing little insight to the problem of question-answering and lacking interpretability as well. In this work, we argue that while video is presented in frame sequence, the visual elements (\eg, objects, actions, activities and events) are not sequential but rather hierarchical in semantic space. To align with the multi-granular essence of linguistic concepts in language queries, we propose to model video as a conditional graph hierarchy which weaves together visual facts of different granularity in a level-wise manner, with the guidance of corresponding textual cues. Despite the simplicity, our extensive experiments demonstrate the superiority of such conditional hierarchical graph architecture, with clear performance improvements over prior methods and also better generalization across different type of questions. Further analyses also demonstrate the model's reliability as it shows meaningful visual-textual evidences for the predicted answers.
\end{abstract}

\begin{figure*}[h]
 \centering
 \scalebox{0.9}{
 \includegraphics[width=1.0\textwidth]{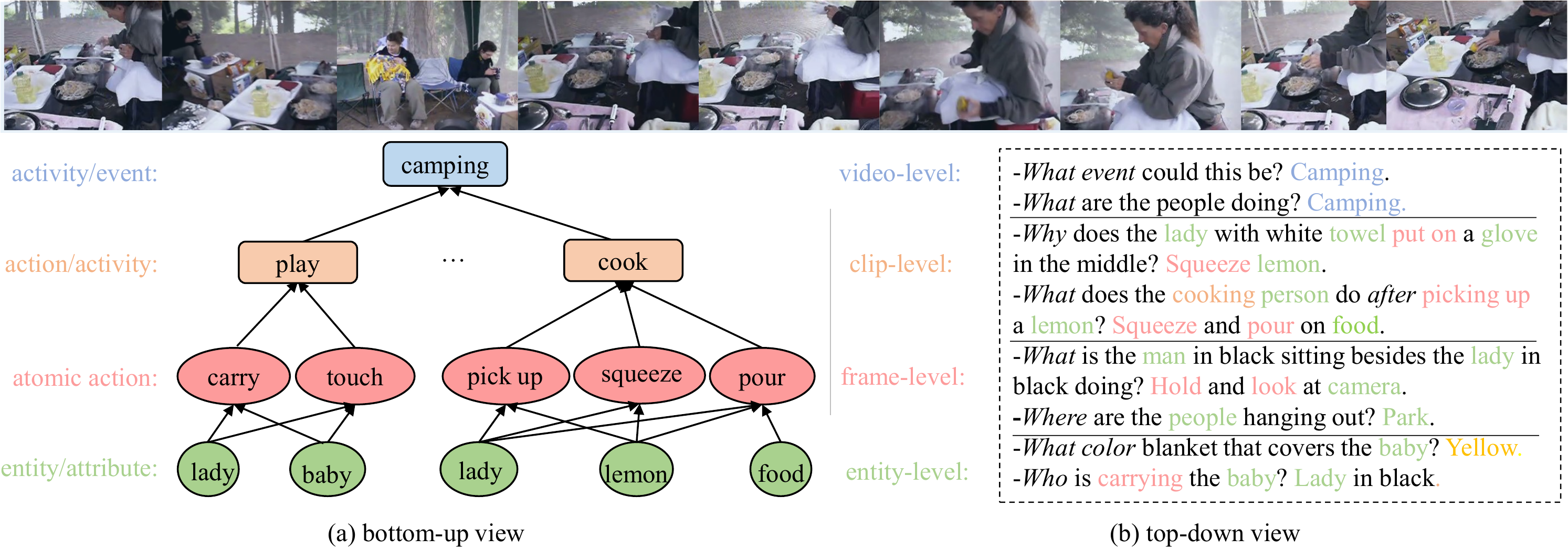}
 }
 \caption{The bottom-up and top-down insights for video question answering. (a) From a bottom-up view, the video contents are hierarchical from low-level visual entities, their local interactions, to high-level activities and global event. (b) From a top-down view, different questions demand video elements at different granularity (\eg, objects, actions, activities and events) for answers, and a single question might also invoke multi-level visual resources to comprehend.}
 \label{fig:intro_butd}
 \vspace{-1.0em}
\end{figure*}

\section{Introduction}
The past few years have witnessed a flourish of research in video-language tasks. Video question answering (VideoQA) is one of the most prominent, given its promise to develop interactive AI and communicate with the dynamic visual world via natural language. Despite its popularity, the challenge in VideoQA remains significant; it demands a wide spectrum of recognitive \cite{ren2015faster,he2016deep,carreira2017quo}, reasoning \cite{hu2018explainable} as well as grounding \cite{hu2017modeling,xiao2020visual} capabilities to comprehend the questions \cite{mao2016generation} and deduce the correct answers. Existing efforts \cite{jang2017,gao2018motion,yu2018joint,fan2019heterogeneous,li2019beyond, jiang2020divide} concentrate on capturing the sequential nature of video frames and question words. As the hierarchical, compositional structure of video contents and the multi-granular essence of linguistic concepts are unaccounted, current models are limited in insight and interpretability, and also their results can be sub-optimal.

In this work, we delve into the problem of video question answering based on a bottom-up and top-down insight\footnote{Bottom-up is from low-level visual contents (video) to high-level semantics (language query), and vice versa for top down.}. As demonstrated in Figure \ref{fig:intro_butd}, from a bottom-up view, video contents are hierarchical in semantic space. Each atomic action (\eg, \texttt{pick up}) involves a visual subject (\eg, \texttt{lady}) and optionally an object (\eg, \texttt{lemon}). Besides, the atomic actions (\eg, \texttt{pick up, squeeze, pour}) compositionally form a super-action or activity (\eg, \texttt{cook}), while activities (\eg, \texttt{cook} and \texttt{play}) further constitute a global event (\eg, \texttt{camping}). From a top-down view, questions are diverse and answering them demands visual information from different granularity levels. Generally, questions concerning objects and their attributes rely on specific visual entities for answers; questions regarding spatial relations and contact actions (\eg, \texttt{hug, kiss, hold, carry}) are better answered based on object interactions at certain frames; while questions about dynamic actions (\eg, \texttt{pick up, put down}), temporal interactions and global event may require aggregating information from multiple video frames or clips. In addition, a single question may also invoke multiple levels of visual elements, which further demands the awareness of the multi-granularity of both video elements and linguistic concepts.

To capture such insight, we propose to model video as a graph hierarchy with the condition of language query in a level-wise fashion. Concretely, the model weaves together visual facts from low-level entities to higher level video elements through graph aggregation and pooling, during which the local and global textual query are incorporated into different levels to match and pinpoint the relevant video elements at the corresponding granularity levels (\eg, objects, actions, activities and events). In this way, our model can not only identify the query-specific visual objects and actions, but also capture their local interactions as well as infer the constituted activities and events. Such versatility is the first of its kind in VideoQA and is of crucial importance in handling diverse questions concerning different video elements. To validate the effectiveness, we test the model on four VideoQA datasets that challenge the various aspects of video understanding and achieve consistently strong results.

To summarize our main contributions: 
1) We provide a bottom-up and top-down insight to advance video question answering in a multi-granular fashion. 2) We propose a hierarchical conditional graph model, which serves as an initial prompt, to capture such insight for VideoQA. 3) Extensive experiments evince that our model is effective and is of enhanced generalizability and interpretability; it achieves the state-of-the-art (SOTA) results across different datasets with various type of questions and finds introspective evidences for the predicted answers as well.

\section{Related Work}
\label{sec:review}
\textbf{Canonical approaches} 
use techniques such as cross-modal attention \cite{jang2017,zeng2017leveraging,li2019beyond,jin2019multi,gao2019structured, jiang2020divide} and motion-appearance memory \cite{xu2017video,gao2018motion,fan2019heterogeneous} to fuse information from the video and question for answer prediction. These methods focus on designing sophisticated cross-modal interactions, whereas treating video and question as a holistic sequence of frames and words respectively. Sequential modelling neither capture the topological (\eg, hierarchical or compositional) structure of the visual elements nor multi-granularity of linguistic concepts. Consequently, the derived QA models are weak in relation reasoning and handling question diversity. 

\textbf{Graph-structured models} \cite{kipf2016semi,velivckovic2018graph} are recently more favoured, 
either for their superior performance in relation reasoning \cite{li2019relation,hu2019language}, or for the improved interpretability \cite{norcliffe2018learning}. L-GCN \cite{huang2020location} constructs a fully-connected graph over all the detected regions in space-time and demonstrates the benefits of utilizing object locations. However, the monolithic graph is cumbersome to extend to long videos with multiple objects. More recently, GMIN \cite{gu2021graph} builds a spatio-temporal graph over object trajectories and shows improvements over its attention version \cite{jin2019multi}. While L-GCN and GMIN construct query-blind graphs, HGA \cite{jiang2020reasoning}, DualVGR \cite{wang2021dualvgr} and B2A \cite{park2021bridge} design query-specific graphs for better performance. Yet, their graphs are built over coarse video segments in a flat way. On the one hand, these models cannot reason fine-grained object interactions in space-time. On the other hand, they are unable to reflect the hierarchical nature of video contents.

\textbf{Hierarchical architectures}. HCRN \cite{le2020hierarchical} designs and stacks conditional relation blocks to capture temporal relations. Nonetheless, it focuses on temporal reasoning of single object actions and models relations with a simple mean-pooling. As a result, it fails to generalize well to the scenarios with multiple object interacted in space-time \cite{xiao2021next}. A very recent work HOSTR \cite{dang2021hierarchical} follows a similar design philosophy to learn a hierarchical video representation, but introduces a nested graph for spatio-temporal reasoning over object trajectories, and achieves better performance. Yet, both HCRN and HOSTR
target at general-purpose visual reasoning. They lack the insights that 1) some video elements (\eg, object, places, spatial relations and contact actions) are easier to identify at frame-level \cite{gkioxari2018detecting,xiao2020visual}, and 2) different parts of a question may invoke visual information at different granularity levels \cite{chen2020fine}. In addition, HOSTR's good performance relies on accurate object trajectories which are hard to obtain in practice, especially for long video sequences with complex object interactions. In this work, we design and emphasize the hierarchical architecture to realize the bottom-up and top-down insights which enables vision-text matching at multi-granularity levels.
\begin{figure*}[t]
 \centering
 \scalebox{0.9}{
 \includegraphics[width=0.8\textwidth, height=0.25\textheight]{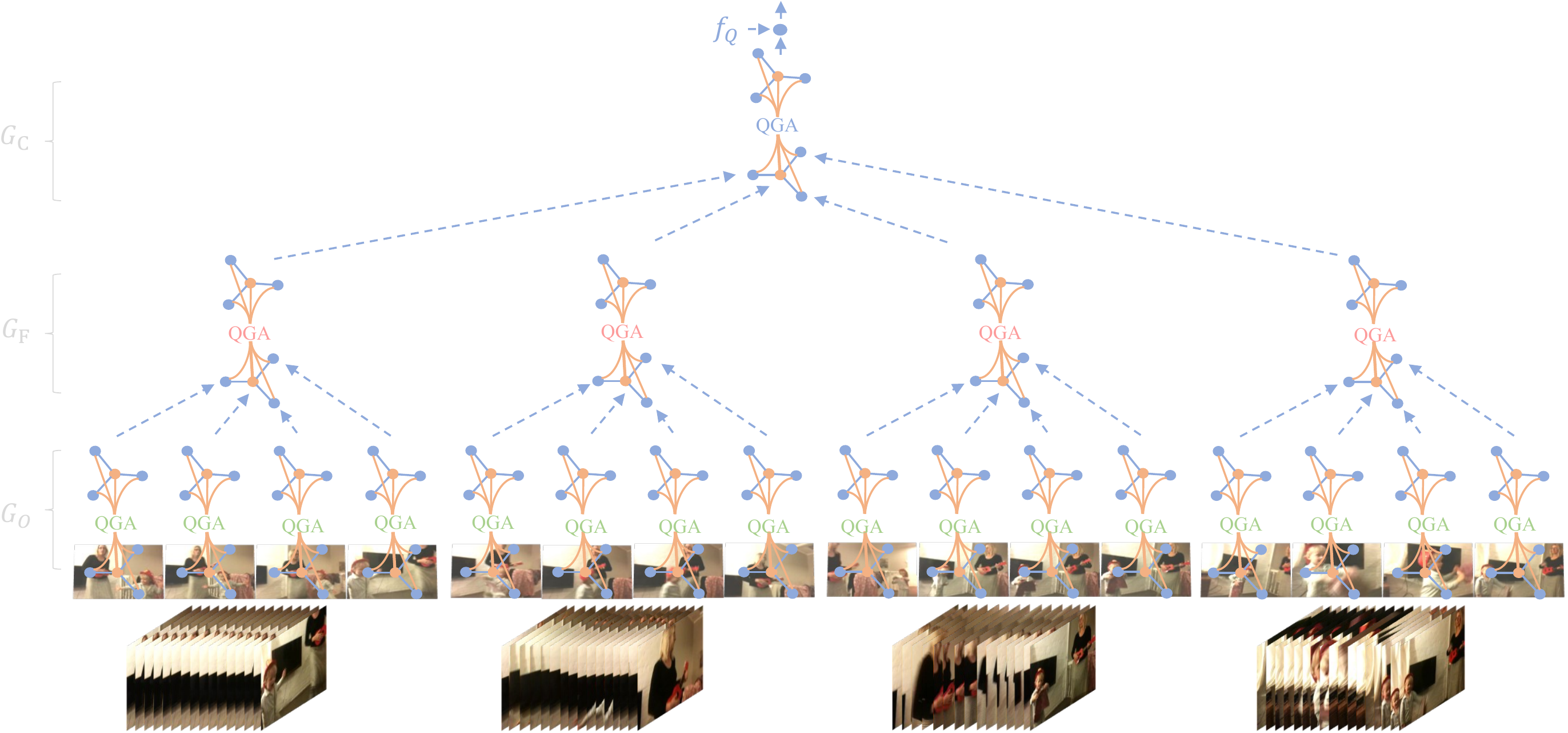}
 }
 \caption{A conditional graph hierarchy built over 4 video clips. $G_O$, $G_F$ and $G_C$ are graphs defined over objects, their interactions across frames and across clips respectively. The graphs are stacked hierarchically to abstract low-level, locally-related visual components into a high-level global representation. The language query $Q$ is conditioned at each level to guide the graph construction. The final aggregated video representation is fused with global query feature $f_Q$ for answer prediction.}

 \label{fig:model_arc}
 \vspace{-1.0em}
\end{figure*}

\section{Method}
\subsubsection{Overview}
Given a video $\mathcal{V}$ and a question $q$, VideoQA aims to predict the correct answer $a^*$ that is relevant to the visual content. Currently, the two typical QA formats are multi-choice QA and open-ended QA. In multi-choice QA, each question is presented with several candidate answers $\mathcal{A}_{mc}$, and the problem is to pick the correct one:
\begin{equation}
    a^* = \arg\max_{a \in \mathcal{A}_{mc}} {\mathcal{F}_{\theta}(a | q, \mathcal{V}, \mathcal{A}_{mc})}.
\end{equation}
In open-ended QA, no answer choices are provided. The task is popularly set as a classification problem to classify the video-question pairs into a globally defined answer set $\mathcal{A}_{oe}$.
\begin{equation}
    a^* = \arg\max_{a \in \mathcal{A}_{oe}} {\mathcal{F}_{\theta}(a | q, \mathcal{V})}.
\end{equation}

To address the problem, we realize the introduced bottom-up and top-down insights by modelling video as a conditional graph hierarchy. As illustrated in Figure \ref{fig:model_arc}, we accomplish this by designing a Query-conditioned Graph Attention unit (\ie, QGA, see Figure~\ref{fig:qga}) and further applying it to reason and aggregate video elements of different granularity levels into a global representation for question answering.

We next elaborate on our model design by first introducing the data representation, then the QGA unit, and finally the hierarchical architecture and answer decoder. 

\subsection{Data Representation}
\textbf{Video.}
We extract a video at $p$ frames per second and then partition it into $K$ clips of length $L$. For each clip, we maintain a dense stream of $L$ frames to obtain the clip-level motion feature and a sparse stream of $\gamma L$ frames ($\gamma \in (0, 1)$) to obtain the region and frame appearance features. In our implementation, the motion features $F_m=\{f_{m_k}\}_{k=1}^K$ and frame appearance features $F_a=\{f_{a_t}\}_{t=1}^T$ ($T=\gamma LK$) are extracted from pre-trained CNNs, specifically 3D version ResNeXt-101 \cite{hara2018can} for motion and ResNet-101 \cite{he2016deep} for frame appearance. Importantly, we also extract $N$ RoIs' (region of interest) appearance features $F_r=\{f_{r_n}\}_{n=1}^N$ along with their bounding boxes from each frame in the sparse stream, using a pre-trained object detector \cite{anderson2018bottom}.

After the extraction, all three types of features are projected into a $d$-dimension space. Specifically, for motion and frame appearance features, the projections are achieved by applying two respective 1-D convolution operations along the time dimension, in which the window sizes are set to 3 to consider the previous and next neighbors as contexts. For clarity, we denote the projected features as $F_m^{K \times d}=\{f_{m_k}^d\}_{k=1}^K$ for motion and $F_a^{T\times d}=\{f_{a_t}^d\}_{t=1}^T$ for frame appearance. For each object, to retain both semantic and geometric information, we jointly represent its RoI appearance $f_r$, bounding box location $f_s$, and temporal position $f_t$ similar to \cite{huang2020location}. The final object feature comes from concatenating the three components and projecting them into $d$-dimensions with a linear transformation followed by an ELU activation: $f_o = \mathrm{ELU}(W_o[f_r;f_s;f_t]),$ in which $[;]$ denotes concatenation and $W_o$ are the parameters of the linear projection. Again, for clarity, we denote the projected object features in a frame as $F_o^{N\times d}=\{f_{o_n}^d\}_{n=1}^N$.

\textbf{Question.}
To obtain a well-contextualized word representation, we extract the token-wise sentence embeddings from the penultimate layer of a fine-tuned BERT model \cite{devlin2018bert} (refer to Appendix B for details). Similar to \cite{xiao2021next}, we further apply a Bi-GRU \cite{cho2014learning} to project the word representations into the $d$-dimension space as the visual part for convenience of cross-modal interaction. Consequently, a language query of length M is represented by 
$
Q^{M \times d}=\{q_m^d|q_m=[\overrightarrow{q_i}; \overleftarrow{q_i}]\}_{m=1}^M,
$
where $\overrightarrow{q_i}^{\frac{d}{2}}$ and $\overleftarrow{q_i}^{\frac{d}{2}}$ denoted the forward and backward hidden states respectively. Particularly, the last hidden state $q_{M}^d$ is treated as the global query representation $f_Q$.

\subsection{Conditional Graph Reasoning and Pooling}
\begin{figure}[t]
\centering
\includegraphics[width=0.3\textwidth]{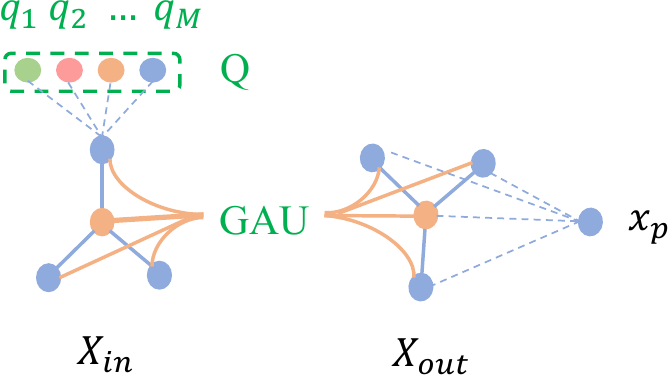}
\caption{Illustration of QGA unit. Color nodes in $Q$ denote the words referring to video elements at different levels.}
\label{fig:qga}
\vspace{-1.0em}
\end{figure}
In this section, we introduce QGA - the key component in our model architecture. As illustrated in Figure \ref{fig:qga}, QGA first contextualizes a set of input visual nodes $X_{\text{in}}$ in relation to their neighbors in both the semantic and geometric (realized by the geometric embeddings of the nodes) space, under the condition of a language query $Q^{M \times d}$, and then aggregates the contextualized output nodes $X_{\text{out}}$ into a single global descriptor $x_p$. The input nodes $X_{\text{in}}^{* \times d}=\{x_{\text{in}_{i}}\}_{i=1}^*$ depend on the hierarchy level; at the bottom, $X_{\text{in}}$ are the object features $F_o$, while at higher levels, the inputs are the outputs of QGAs at the preceding level. The dimension varies accordingly and we use * as a placeholder for the number of input nodes.

\textbf{Query Condition.} By condition, we pay attention to the video elements that are invoked in the questions, which is achieved by augmenting the corresponding nodes' representations with $Q$:

\begin{equation}
\hat{x} = x_{\text{in}} + \sum\nolimits_{j=1}^M \alpha_m q_m, \quad \text{where} \quad \alpha = \sigma(x_{\text{in}}Q^\mathrm{'}), 
\end{equation}
where $\sigma(\cdot)$ is the softmax normalization function and $\mathrm{'}$ indicates matrix transpose. The indices of $x$ are omitted for brevity. We expect that, through the cross-attention aggregation of the word representations with respect to the visual node $x_{in}$, the visual node's textual correspondence will have a stronger response in the aggregation if the represented video element is mentioned in the question. As a result, the corresponding video element (represented by $x_{\text{in}}$) will be highlighted and contribute more to subsequent operations.

\textbf{Graph Attention.}
After obtaining the augmented node representations $\hat{X}^{*\times d}=\{\hat{x}\}_{i=1}^*$, the edges (including self-loops) represented by the values of the adjacency matrix $A^{*\times *}$ are dynamically computed as the similarities between the node pairs:

\begin{equation}
    A= \sigma(\phi_{W_{av}}(\hat{X})\phi_{W_{ak}}(\hat{X})^\mathrm{'}), 
\end{equation}
in which the function $\phi$ denotes linear transformation with learnable parameters $W_{av}$ and $W_{ak} \in \mathbb{R}^{d \times \frac{d}{2}}$ respectively. The $\text{softmax}$ operation normalizes each row, so that the $i^{th}$ row $A_i$ denotes the attention values (\ie, values of normalized dot-product) of node $i$ with regard to all the other nodes. We then apply a $H$-layer graph attention aggregation with skip-connections to refine the nodes in relation to their neighbors based on the adjacency matrix $A$: 
\begin{equation}
    \hat{X}^{(h)} = \mathrm{ReLU}((A+I)\hat{X}^{(h-1)}W^{(h)}), 
\end{equation}
where $W^{(h)} \in \mathbb{R}^{d\times d}$ and $\hat{X}^{(h)}$ are the parameters and outputs of the $h^{th}$-layer graph attention respectively. $\hat{X}^{(0)}$ is initialized with the query-attended node $\hat{X}$. $I$ is the identity matrix for skip connections. The final output is obtained by a last skip-connection: $X_{\text{out}} = \hat{X} + \hat{X}^{(H)}$.

\textbf{Node Aggregation.} To get an aggregated representation $x_p$ for a QGA unit, we apply self-attention pooling \cite{lee2019self} over the set of output nodes:
\begin{equation}
   x_p = \sum\nolimits_{i=1}^* \beta_i x_{\text{out}_i}, \quad \text{where} \quad \beta = \sigma(\phi_{W_p}(X_{\text{out}})),
\end{equation}
where $W_p \in \mathbb{R}^{d \times 1}$ are the learnable linear mapping weights. 

\subsection{Hierarchical Architecture}
In this section, we explain how to apply the QGA units to achieve the hierarchical architecture to reflect the bottom-up and top-down insights for question answering. As shown in Figure \ref{fig:model_arc}, the QGA units at the bottom level ($G_O$) take as inputs a set of object features frame-wisely and capture a static picture of object interactions:
\begin{equation}
F_{G_O} = G_O(F_O) = \text{QGA}(F_O), 
\end{equation}
The output features $F_{G_O}^{T\times d}=\{f_{{G_O}_t}\}_{t=1}^T$ are then combined with the frame appearance features $F_a^{T \times d}$ (serve as global context) by concatenation:
\begin{equation}
\label{equ:merge}
\bar{f}_{{G_O}_t} = \text{ELU}(W_{ao}[f_{a_t}; f_{{G_O}_t}]), \quad t\in \{1, 2, \cdots, T\},
\end{equation}
where $W_{ao} \in \mathbb{R}^{2d \times d}$ are linear parameters. Then, the QGA units at the second level ($G_F$) are applied to $\bar{F}_{G_{O}}^{T\times d}=\{\bar{f}_{{G_O}_t}\}_{t=1}^T$ clip-wisely to model a short-term interaction dynamics, as well as to reason and aggregate the low-level visual components to a higher granularity level (\eg, from actions to activities):
\begin{equation}
F_{G_F} = G_F(\bar{F}_{G_O}) = \text{QGA}(\bar{F}_{G_O}).   
\end{equation}
\noindent The outputs features $F_{G_F} \in \mathbb{R}^{K\times d}$ are then combined with the global motion feature $F_m^{K \times d}$ to obtain $\bar{F}_{G_F}$ in a way analogous to Equ.~\ref{equ:merge}. Furthermore, the QGA unit at the top level ($G_C$) operates over $\bar{F}_{G_F}$ to reason about the local, short-term interactions, and aggregate them into a single global representation $f_V$ over the entire video:
\begin{equation}
f_V = G_C(\bar{F}_{G_F}) = \text{QGA}(\bar{F}_{G_F}).   
\end{equation}

Overall, the hierarchical architecture is achieved by nesting the conditional graph operations that can be conceptually represented as
\begin{equation}
    f_V=G_C(G_F(F_a, G_O(F_o)), F_m).
\end{equation}
\noindent Finally, $f_V \in \mathbb{R}^d$ is passed to the answer-decoder, along with the global query representation $f_Q$, to jointly determine the correct answers.

By the hierarchical graph structure, the video elements of different granularity can be level-wisely inferred. Besides, by the multi-level token-level query conditions, the model is capable of pinpointing the referred video elements at different granularity, and it is also flexible in handling different textual queries. In addition, by introducing the context features (\ie, $F_a$ and $F_m$), the model can make up the downside of lacking the respective global information at each level. Importantly, the model is of enhanced flexibility and interpretability, from a perspective of query-instantiated neural modular networks \cite{hu2018explainable} and from a introspective analysis of the learned attention weights as our model are purely attention-based, respectively.

\subsection{Answer Decoder}
For multi-choice QA, we concatenate each candidate answer with the corresponding question to form a holistic query. The resulting global query feature $f_Q$ is fused with the final video feature $f_V$ via Hadamard product (\aka, element-wise product) $\odot$. Then, a full-connected layer with softmax is applied as classifier: 
\begin{equation}
    s = \sigma(W_c^\mathrm{'}(f_Q \odot f_V)+b),
\end{equation}
where $W_c \in \mathbb{R}^{d\times 1}$ and $b$ are learnable parameters. s is the prediction score. During training, we maximize the margin between the positive and negative QA-pairs (\ie, $s^p$ and $s^n$ respectively) with the hinge loss:
    $\mathcal{L}=\sum\nolimits_{i=1}^{|\mathcal{A}_{mc}|} \max(0, 1+s_i^n-s^p)$,
where $|\mathcal{A}_{mc}|$ is the number of choices in a question.

For open-ended QA, as the number of categories (answers) are large, we empirically find that it is better to concatenate the global question feature $f_Q$ with the video feature $f_V$ before the classifier.
\begin{align}
    s = \sigma(W_c^\mathrm{'}(\mathrm{ELU}(W_{qv}[f_Q; f_V]))+b),
\end{align}
in which $W_{qv}\in \mathbb{R}^{2d\times d}$, $W_c \in \mathbb{R}^{d \times |\mathcal{A}_{oe}|}$, and $b^{|\mathcal{A}_{oe}|}$ are learnable parameters. Besides, $|\mathcal{A}_{oe}|$ is the size of the predefined answer set. During training, the optimization is achieved by minimizing the cross-entropy loss:
   $ \mathcal{L}= -\sum\nolimits_{i=1}^{|\mathcal{A}_{oe}|} y_i \log s_i$,
where $s_i$ is the prediction score for the $i^{th}$ sample. $y_i=1$ if the answer index corresponds to the $i^{th}$ sample's ground-truth answer and 0 otherwise.

\section{Experiments}
\subsection{Datasets}
We experiment on four VideoQA datasets that challenge the various aspects of video understanding:
\textbf{TGIF-QA}~\cite{jang2019video} features questions about action repetition, state transition and frame QA. Action repetition includes the sub-tasks of repetition counting and repeating action recognition. In this work, we experiment on the latter since the prediction of numbers are hard to explain. 
\textbf{MSRVTT-QA} and \textbf{MSVD-QA} mainly challenge a recognition of video elements. Their question-answer pairs are automatically generated from the respective video descriptions by \cite{xu2017video}. \textbf{NExT-QA} \cite{xiao2021next} is a challenging benchmark that goes beyond superficial video description to emphasize causal and temporal multi-object interactions. It is rich in object relations in space-time \cite{shang2019annotating}. The dataset has both multi-choice QA and generation-based QA. In this work, we focus on the former and leave the generation-based QA for future exploration. For all datasets, we report accuracy (percentage of correctly answered questions) as the evaluation metric. Other statistical details are given in Appendix A.

\setlength{\belowcaptionskip}{-0.2cm}
\begin{table*}[t!]
\small
\centering
\begin{threeparttable}
\scalebox{0.9}{
\begin{tabular}{l @{\hskip 0.2in} c @{\hskip 0.2in} c @{\hskip 0.2in} c @{\hskip 0.2in} c @{\hskip 0.2in} |c @{\hskip 0.2in} c @{\hskip 0.2in} c @{\hskip 0.2in}  c@{\hskip 0.2in}}
\specialrule{.1em}{.05em}{.05em} 
   \multirow{2}*{Models} & \multicolumn{4}{c}{NExT-QA Val} & \multicolumn{4}{c}{NExT-QA Test} \cr
& Causal & Temporal & Descriptive & Overall & Causal & Temporal & Descriptive & Overall \cr
\specialrule{.1em}{.05em}{.05em}
    ST-VQA & 44.76   & 49.26   & 55.86   & 47.94 & 45.51   & 47.57   & 54.59   & 47.64 \cr
    Co-Mem & 45.22   & 49.07  & 55.34  & 48.04  & 45.85   & \underline{50.02}  & 54.38  & 48.54 \cr
    HME  & 46.18   & 48.20   & 58.30  & 48.72   & 46.76   & 48.89   & 57.37  & 49.16 \cr
    L-GCN & 45.15   & 50.37   & 55.98  & 48.52 & 47.85   & 48.74   & 56.51  & 49.54  \cr
    HGA  & \underline{46.26}  & \underline{50.74}  & \underline{59.33}  & \underline{49.74}  & \underline{48.13}  & 49.08  & \underline{57.79}  & \underline{50.01} \cr
    HCRN & 45.91   & 49.26   & 53.67  & 48.20  & 47.07   & 49.27   & 54.02  & 48.89 \cr
    \textbf{HQGA (Ours)} & \textbf{48.48}  & \textbf{51.24}  & \textbf{61.65}  & \textbf{51.42}  & \textbf{49.04}  & \textbf{52.28}  & \textbf{59.43}  & \textbf{51.75} \cr
\specialrule{.1em}{.05em}{.05em} 
\end{tabular}
}
\caption{Comparison of accuracy. The best and second-best results are highlighted in bold and underline respectively.}
\label{table:next_result}
\end{threeparttable}
\vspace{-1.0em}
\end{table*}

\subsection{Implementation Details}
We extract each video in NExT-QA, MSVD-QA and MSRVTT-QA at $p=\{6, 15, 15\}$ frames per second respectively. For TGIF-QA, all the frames are used. Based on the average video lengths, we uniformly sample $K=\{16, 8, 8, 4\}$ clips for each video in the four datasets respectively, while fixing the clip length $L=16$. The sparse stream is obtained by evenly sampling with $\gamma=0.25$. For each frame in the sparse stream, we detect $N=20$ regions for NExT-QA and 10 for the others. The dimension of the models' hidden states is $d=512$ and the default number of graph layers in QGA is $H=2$. For training, we adopt a two-stage scheme by firstly training the model with learning rate $lr=10^{-4}$ and then fine-tune the best model obtained in the $1st$ stage with a smaller $lr$, \eg, $5\times10^{-5}$. For both stages, we train the models by using Adam optimizer with batch size of 64 and maximum epoch of 25. Other details are presented in Appendix B.

\subsection{The State of the Art Comparison}
In Table \ref{table:next_result} and Table \ref{table:gif_result}, we compare our model with some established VideoQA techniques covering 4 major categories: \textbf{1)} cross-attention (\eg, ST-VQA \cite{jang2017}, PSAC \cite{li2019beyond}, STA \cite{gao2019structured}, MIN \cite{jin2019multi} and QueST \cite{jiang2020divide}), \textbf{2)} motion-appearance memory (\eg, AMU \cite{xu2017video}, Co-Mem \cite{gao2018motion} and HME \cite{fan2019heterogeneous}), \textbf{3)} graph-structured models (\eg, L-GCN\footnote{L-GCN's results on NExT-QA and MSRVTT-QA are reproduced by us with the official code.} \cite{huang2020location}, HGA\cite{jiang2020reasoning}, DualVGR \cite{wang2021dualvgr}, GMIN \cite{gu2021graph} and B2A \cite{park2021bridge}) and \textbf{4)} hierarchical models (\eg, HCRN \cite{le2020hierarchical} and HOSTR \cite{dang2021hierarchical}). The results show that our Hierarchical QGA (HQGA) model performs consistently better than the others on all the experimented datasets. 


Particularly, both L-GCN and GMIN are graph-based methods that focus on leveraging object-level information (similar to us) for question-answering. However, they model the object either in a monolithic way in space-time or in trajectories. This neither reflect the hierarchical nor the compositional nature of the video elements. Furthermore, their graphs are constructed without the guidance of language queries. By filling such gaps, our model shows clear superiority to both methods on the experimented datasets.
\begin{table}[t!]
\small
\centering
\begin{threeparttable}
\scalebox{0.9}{
\begin{tabular}{l @{\hskip 0.1in} c @{\hskip 0.1in} c @{\hskip 0.1in} c @{\hskip 0.1in} | @{\hskip 0.1in}c | @{\hskip 0.1in}c}
\specialrule{.1em}{.05em}{.05em} 
   \multirow{2}*{Models}
    ~ &\multicolumn{3}{c}{TGIF-QA} & MSRV & MSVD \cr
    ~ & Action & Transition & FrameQA & TT-QA & -QA \cr
\specialrule{.1em}{.05em}{.05em} 
    ST-VQA & 62.9 & 69.4 & 49.50 & 30.9 & 31.3 \cr
    PSAC   & 70.4 & 76.9 & 55.7 & - &  -\cr
    STA    & 72.3 & 79.0 & 56.6 & -  & - \cr
    MIN    & 72.7 & 80.9 & 57.1 & 35.4 & 35.0 \cr
    QueST  & \underline{75.9} & 81.0 & \underline{59.7} & 34.6 & 36.1 \cr
\specialrule{.1em}{.05em}{.05em} 
    AMU    & - & - & - & 32.5 & 32.0 \cr
    Co-Mem & 68.2 & 74.3 & 51.5 & 31.9 & 31.7\cr
    HME    & 73.9 & 77.8 & 53.8 & 33.0 & 33.7\cr
\specialrule{.1em}{.05em}{.05em} 
    L-GCN  & 74.3 & 81.1 & 56.3 & 33.7 & 34.3 \cr
    HGA    & 75.4 & 81.0 & 55.1 & 35.5 & 34.7 \cr
    DualVGR & - & - & - & 35.5 & 39.0 \cr
    GMIN   & 73.0 & 81.7 & 57.5 & 36.1 & 35.4 \cr
    B2A    & \underline{75.9} & 82.6 & 57.5 & \underline{36.9} & 37.2  \cr
\specialrule{.1em}{.05em}{.05em} 
    HCRN   & 75.0 & 81.4 & 55.9 & 35.6 & 36.1 \cr
    HOSTR & 75.0 & \underline{83.0} & 58.0 & 35.9 & \underline{39.4} \cr
    \textbf{HQGA}  & \textbf{76.9}  & \textbf{85.6}  & \textbf{61.3}  & \textbf{38.6} & \textbf{41.2} \cr
\specialrule{.1em}{.05em}{.05em} 
\end{tabular}
}
\caption{Comparison of accuracy.}
\label{table:gif_result}
\end{threeparttable}
\vspace{-1.0em}
\end{table}

HCRN and HOSTR are similar to us in designing hierarchical conditional architectures. Nonetheless, HCRN is limited to hierarchical temporal relations between frames, in which the relations are modeled by simple average pooling. It is helpful for identifying repeated actions and state transition of single object (see results on TGIF-QA), but it is insufficient to understand more complicated object interactions in space-time. As a result, it performs even worse than L-GCN on NExT-QA. HOSTR advances HCRN by building the hierarchy over object trajectories and adopting graph operation for relation reasoning. Yet still, it focuses merely on designing general-purpose neural building blocks and lacks the bottom-up and top-down insight (Figure \ref{fig:intro_butd}) for VideoQA, which results in its sub-optimal model design and results.

While HCRN and HOSTR use global query representations, QueST breaks down the question into spatial and temporal components, and designs separated attention modules to aggregate information from video for question answering. It obtains good results on TGIF-QA but does not generalize well to MSRVTT-QA and MSVD-QA where the questions do not have such spatial and temporal syntactic structure. Finally, the methods HGA, DualVGR and B2A, like us, try to align words in the language query with their visual correspondences in the video. However, all of them adopt a flat way of alignment at segment level and lack the hierarchical structure, which most likely accounts for their inferiority.

\begin{figure*}[t]
 \centering
 \scalebox{0.9}{
  \includegraphics[width=1.0\textwidth]{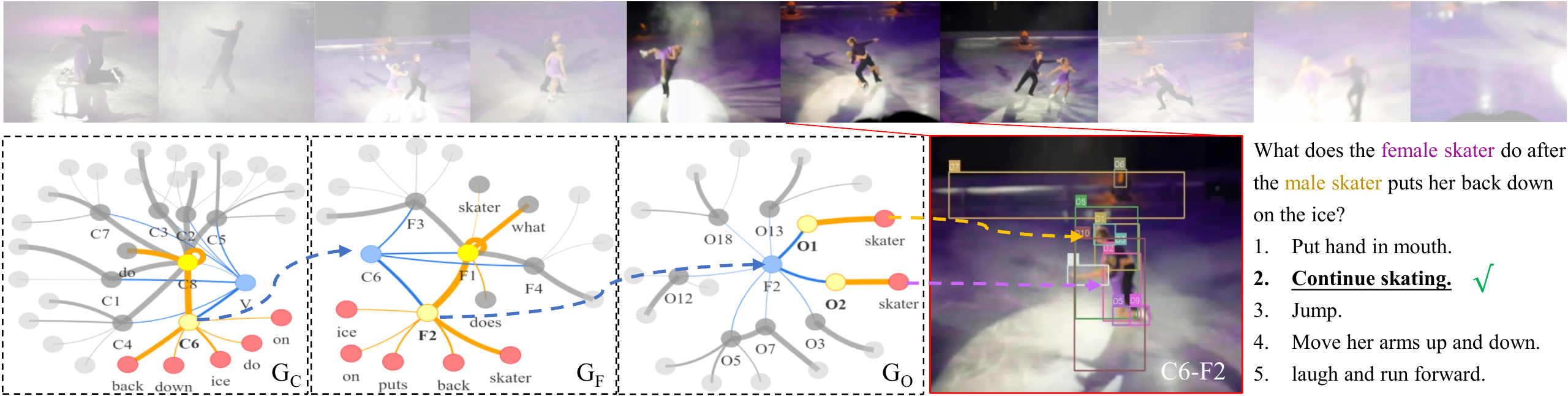}
 }
 \caption{A correct prediction case from NExT-QA \cite{xiao2021next}. $G_C$, $G_F$, $G_O$ show the learned graph and conditional attention weights at different levels. The nodes and edges that response relatively stronger to the aggregated nodes are highlighted. (Blue: weights of self-attention pooling $\beta$. Orange: weights of adjacency matrix $A$ and query condition $\alpha$.)}
 \label{fig:qual_result}
\vspace{-.5em}
 \end{figure*}

\subsection{Model Analysis}
We analyze our model on the validation sets of NExT-QA and MSRVTT-QA. We first conduct an ablation study on the number of graph attention layers $H$, sampled video clips $K$ and regions $N$ to find the optimal settings on the two datasets (refer to Appendix C for more details). Then, we fix the optimal settings for subsequent experiments.

\textbf{Hierarchy.} The top section of Table \ref{table:aba_results} shows that accuracy drops by 0.9\% on both datasets when removing the QGA units at the bottom ($w/o$ $G_O$) ($F_a$ and $F_m$ are used as respective inputs for $G_F$ and $G_C$.). The results demonstrate the importance of modeling the static picture of object interactions. Taking away the $2nd$ level graph units ($w/o$ $G_F$) by directly applying $G_C$ over $G_O$'s outputs corresponding to the respective middle frames of the clips, we can observe even more drastic accuracy drops (over 1.2\%) on both datasets. This results demonstrate the critical role of $G_F$ in modeling the short-term dynamics of object interactions. Finally, when we remove both $G_O$ and $G_F$ and apply a monolithic graph over the clips (directly using $F_m$ as inputs for $G_C$), the accuracy drops sharply by 1.5\% and 2.6\% on NExT-QA and MSRVTT-QA respectively. This clearly shows that only modeling clip-level information 
is unsatisfactory. The comparisons confirm the significance of hierarchically weaving together video elements of different levels.

\textbf{Graph.} The middle section of Table \ref{table:aba_results} shows that substituting the top QGA ($G_C$) with a sum-pooling over the corresponding input nodes ($w/o$ $G_C(s)$), 
leads to accuracy drops from HQGA of 0.68\% and 0.54\% on NExT-QA and MSRVTT-QA respectively. The result validates the importance of $G_C$ in reasoning over local, short-term dynamic interactions. Replacing the middle QGAs ($G_F$) with sum-poolings ($(ss)$) further degrades the performance on both datasets. 
The result indicates that a sum-pooling is neither sufficient in capturing the short-term dynamics of object interactions nor capable of relation reasoning over them. Thus, it evinces the strengths of both $G_C$ and $G_F$. Finally, by replacing the last (bottom) QGAs ($G_O$) with sum-poolings ($(sss)$), we can observe a further exacerbation of performances on the basis of the previous two ablations. The results demonstrate $G_O$'s superiority in capturing object relations in static frames. This experiment demonstrates the advantage of graph attention over a simple sum-pooling.

\textbf{Multi-level Condition.} 
As shown in Table \ref{table:aba_results} (bottom part), the results in the first three rows show that removing the language condition at any single level jeopardize the overall performance, indicating the necessity of injecting the query cues at multiple levels. Specially, we investigate replacing the token-wise query representations $Q$ with a global one $f_Q$, \ie, by concatenating $f_Q$ with the respective graph nodes at all levels. From the results shown in the row $w/f_Q$, we find that a global condition can slightly boost the performance on MSRVTT-QA compared with the model variant without any conditioning (\eg, 37.52\% \vs 37.03\%). However, such benefit disappears when the model is extrapolated to the scenario where the videos and questions are much more complex (\eg, NExT-QA). Finally, we conduct additional ablation studies on $F_a$ and $F_m$ that serve as global contexts to enhance the representations of graph nodes at the corresponding levels. The results show that both features help a bit to the overall performance.

\textbf{Discussion.}
By jointly considering the ablation results of graph hierarchy and multi-level condition in Table \ref{table:aba_results}, we can see that, compared with the graph hierarchy, the multi-level query condition has relatively smaller influence on the performance. We speculate that some questions are too simple to provide meaningful referring clues to the video contents (\eg, \texttt{`what is happening'}). Such questions purely rely the model to reason the video contents for answers. 
\begin{table}[t!]
\small
\centering
\begin{threeparttable}
\scalebox{0.9}{
\begin{tabular}{l @{\hskip 0.4in} c @{\hskip 0.1in} c @{\hskip 0.1in}}
\specialrule{.1em}{.05em}{.05em} 
    Model Variants
     & NExT-QA & MSRVTT-QA \cr
\specialrule{.1em}{.05em}{.05em}
    \textbf{HQGA} &  \textbf{51.42} & \textbf{38.23} \cr
    w/o $G_O$ &  50.50 & 37.26 \cr
    w/o $G_F$ &  50.00 & 37.05 \cr
    w/o $G_O$ \& $G_F$ & 49.96 & 35.66 \cr
\specialrule{.1em}{.05em}{.05em} 
    w/o $G_C$(s) &  50.74 & 37.69 \cr
    w/o $G_C$ \& $G_F$(ss) & 50.44 & 36.94 \cr
    w/o $G_C$ \& $G_F$ \& $G_O$(sss)  & 50.32 & 35.88 \cr
\specialrule{.1em}{.05em}{.05em}
    w/o $Q_C$  & 51.30 & 38.17\cr
    w/o $Q_C$ \& $Q_F$ & 51.08 & 37.62 \cr
    w/o $Q_C$ \& $Q_F$ \& $Q_O$  & 50.62 & 37.03 \cr
    w/  $f_Q$  & 50.16 & 37.52 \cr
    w/o $F_m$  & 50.90 & 37.94 \cr
    w/o $F_a$ \& $F_m$  & 50.34 & 37.86\cr
\specialrule{.1em}{.05em}{.05em}
\end{tabular}
}
\caption{Model ablation results on the validation sets.}
\label{table:aba_results}
\end{threeparttable}
\vspace{-1.5em}
\end{table}

Another finding is that, both the graph hierarchy and multi-level condition have relatively smaller performance gains on NExT-QA than on MSRVTT-QA. A possible reason could be that NExT-QA emphasizes causal and temporal action relations; it requires high-quality action recognition. Yet recognizing actions in such complex multiple-object scenarios remains a significant challenge in video understanding \cite{gu2018ava,feichtenhofer2019slowfast}. Our model can reason on the relations between video elements at multi-granularity levels. However, such capability seriously relies on object appearance features in its current version; the absence of region-level motion is a limitation. Our further analyses of model performances per question type on MSRVTT-QA and MSVD-QA (see Table \ref{table:detailres}) show that there are still clear gaps between action/activity recognition and object/attribute recognition in videos; the gaps remain 4.6\% on MSRVTT-QA and 7.5\% on MSVD-QA for 'what' questions. A promising solution would be to jointly model both the appearance and motion for each object. As it is not the focus of this work, we leave it for future exploration. 

Intriguingly, the results of our simplest model variant ($w/o$ $G_O\&G_F$) are still on par with some previous SOTAs. Such strong results can be attributed to our better data representation. Here to verify BERT, we additionally explore substituting BERT with GloVe \cite{pennington2014glove} and achieve 37.2\% on MSRVTT-QA test set. This result confirms the advantages of BERT (38.6\% \vs 37.2\%) as a good contextualized representation to fulfill the multi-granular condition, \eg, the disambiguation between the male and female skaters denoted as $O_1$ and $O_2$ respectively in Figure  \ref{fig:qual_result} ($G_O$). Also, the result prompts future exploration of finetuning pre-trained vision-text architectures \cite{lei2021less} for potential improvement of performance.

\setlength{\tabcolsep}{2.6pt}
\begin{table}[t!]
\small
\centering
\begin{threeparttable}
\scalebox{1.0}{
\begin{tabular}{lccccccc|c}
\specialrule{.1em}{.05em}{.05em} 
     Datasets & what$_a$ & what$_o$ & what & who & how & when & where & all \cr
\specialrule{.1em}{.05em}{.05em} 
    MSRVTT   & 30.1 & 34.7 & 32.5 & 48.9 & 81.5 & 78.3 & 38.4 & 38.6 \cr
    MSVD     & 25.4 & 32.9 & 30.4 & 57.2 & 76.2 & 75.9 & 32.1 & 41.2 \cr
\specialrule{.1em}{.05em}{.05em} 
\end{tabular}
}
\caption{Test accuracy per question type. Roughly $\frac{1}{2}$ ($\frac{1}{3}$) of the `what' questions in MSRVTT-QA (MSVD-QA) ask actions/activities; others are about objects/attributes. We distinguish them via the pattern `what ... doing' in questions.}
\label{table:detailres}
\end{threeparttable}
\vspace{-1em} 
\end{table}

\textbf{Qualitative Analysis.}
We show a prediction case in Figure \ref{fig:qual_result} (find more examples in Appendix C). Firstly, by tracing down the self-attention pooling weights $\beta$, our model precisely finds the video contents that are relevant to the textual query, \eg, from the $6th$ video clip $C_6$ to its $2nd$ frame $F_2$, and further to the male and female skaters (denoted as $O_1$ and $O_2$ respectively in $F_2$). Secondly, according to the query-conditional weights $\alpha$, our model successfully differentiates the information of different granularity levels for both the video and question contents, and discriminatingly match them at the corresponding levels. For example, the nodes at high level $G_C$ show stronger responses to the action-related words (`\texttt{do}' and `\texttt{back down}'), whereas those nodes at the lower levels ($G_F$ and $G_O$) respond strongly to the visual objects (`\texttt{skater}'). Finally, according to the learned adjacency matrices,  while we construct fully-connected graphs in QGA, the learned connections are quite sparse, suggesting that our model can learn to filter out meaningless relations with respect to the query. 

\section{Conclusion}
This work delves into video question answering and uncovers the insights of bottom-up and top-down for video-language alignment. To capture the insight, we propose to build video as a conditional graph hierarchy which level wisely reasons and aggregates low level visual resources into high level video elements, in which the language queries are injected into different levels to match and pinpoint the video elements at multi-granularity. To accomplish this, we design a reusable query-conditioned graph attention unit and stack it to achieve the hierarchical architecture. Our extensive experiments and analyses have validated the effectiveness of the proposed method. Future attempts can be made on incorporating object-level motion information, or exploiting pretraining techniques to boost the performance.

\section*{Acknowledgements}
This research is supported by the Sea-NExT Joint Lab.
\bigskip
\bibliography{aaai21}
\section*{Appendix}
\section{A. Datasets}
\setlength{\tabcolsep}{7pt}
\begin{table*}[t!]
\small
\centering
\begin{threeparttable}
    \scalebox{1.0}{
    \begin{tabular}{llcccccc}
        \specialrule{.1em}{.05em}{.05em} 
        Datasets & Main Challenges & \#Videos/\#QAs & Train & Val & Test & VLen (s) & QA \cr
        \specialrule{.1em}{.05em}{.05em} 
         MSRVTT-QA & Object \& Action Recognition  & 10K/ 244K & 6.5K/159K & 0.5K/12K & 3K/73K & 15& OE\cr
         \cline{1-8}
          MSVD-QA & Object \& Action Recognition  & 1.97K/ 50K & 1.2K/30.9K & 0.25K/6.4K & 0.52K/13K & 10& OE\cr
         \cline{1-8}
         \multirow{3}*{TGIF-QA} 
         & Repetition Action  & 22.8K/22.7K & 20.5K/20.5K & -& 2.3K/2.3K & 3 & MC\cr
         & State Transition  & 29.5K/58.9K & 26.4K/52.7K & -& 3.1K/6.2K & 3 & MC\cr
         & Frame QA         & 39.5K/53.1K & 32.3K/39.4K & -& 7.1K/13.7K & 3 & OE\cr
         \cline{1-8}
         NExT-QA & Causal \& Temporal Interaction  & 5.4K/48K & 3.8K/34K & 0.6K/5K & 1K/9K & 44 & MC \cr
        \specialrule{.1em}{.05em}{.05em}
    \end{tabular}
    }
    \caption{Details of the datasets for experiment. OE: open ended. MC: multiple choice.}
\label{tab:dataset}
\end{threeparttable}
\end{table*}

Table \ref{tab:dataset} presents the details of the experimented datasets. The four datasets challenge the various aspects of video understanding from simple object/action recognition, state transition, to deeper causal and temporal action interaction among multiple objects. For NExT-QA \cite{xiao2021next}, MSRVTT-QA \cite{xu2017video} and MSVD-QA \cite{xu2017video}, we follow the official training, validation and testing splits for experiments. For each sub-task in TGIF-QA \cite{jang2017}, we randomly split 10\% data from training set to determine the optimal iteration epochs since no validation set is officially provided; then, the whole training set is utilized to learn the final model. 

\begin{table*}[t!]
\small
\centering
\begin{threeparttable}
\scalebox{1.0}{
\begin{tabular}{l @{\hskip 0.1in} c @{\hskip 0.1in} c @{\hskip 0.1in} c @{\hskip 0.1in} | @{\hskip 0.1in}c | @{\hskip 0.1in}c | @{\hskip 0.1in}c}
\hline
   \multirow{2}*{}
     &\multicolumn{3}{c}{TGIF-QA} & MSRVTT-QA & NExT-QA & MSVD-QA\cr
     & Action & Trans. & FrameQA & & \cr
\hline
    QA(Train/Test) & 20.5K/2.3K & 52.7K/6.2K & 39.4K/13.7K & 159K/73K & 38K/10K & 30.9K/13K \cr
    Model Size & 46M & 46M & 52M & 56M & 46M & 50M \cr
    Train Time (M) & $<$ 4  & $<$ 11  & $<$3 & $<$ 15 & $<$ 15 & $<$3 \cr
    Test Time (M) & $<$ 1 & $<$ 1 & $<$ 1 & $<$ 6 & $<$ 4 & $<$ 1 \cr
\hline
\end{tabular}
}
\caption{Model efficiency of HQGA. GPU: Tesla V100. Batch size: 64. M: Minutes per epoch. Time of feature extraction is not considered. The model size varies a bit on open-ended QA datasets as the number of answers (categories) are different.}
\label{tab:model}
\end{threeparttable}
\end{table*}

\section{B. Implementation Details}
The motion features $F_m$ for the dense stream are extracted from a 3D-version ResNeXt-101 model provided by \cite{hara2018can}, and is pre-trained on Kinetics \cite{kay2017kinetics}. For each frame in the sparse stream, the appearance feature $F_a$ are extracted from a ResNet-101 \cite{he2016deep} model pre-trained on ImageNet \cite{deng2009imagenet}. Besides, the object detection model is adopted from \cite{anderson2018bottom}, and is pre-trained on Visual Genome \cite{krishna2017visual} with both objects and attributes. 

For question-answers in NExT-QA, we use the provided official BERT \cite{devlin2018bert} features. For other multi-choice QA (MC) tasks, we concatenate the provided choices with the corresponding question and fine-tune the BERT-base model by maximizing the probability of the correct QA-pair with regard to those negative ones. For open-ended QA (OE), we treat each question as a sentence and the corresponding answer as the category of that sentence, so as to convert the QA task to the problem of sentence classification. We basically fine-tune the models by maximal 3 epochs according to the performance on the respective validation sets. The fine-tuning procedure takes less than half an hour with Tesla V100. Besides, the maximal sentence length is set to 20 for all datasets except for NExT-QA which is 37. 

For both BERT fine-tuning and our model training, the top 4,000 answers of high frequency, plus one anonymous class for answers that are out-of-set, are used as global answer set for MSRVTT-QA. In terms of MSVD-QA and the FrameQA sub-task in TGIF-QA, all of the answers in the respective training sets are treated as the pre-defined answer sets. During evaluation, predictions of out-of-set answers are regarded as failing cases.  
\begin{figure}[t!]
 \centering
 \includegraphics[width=0.47\textwidth]{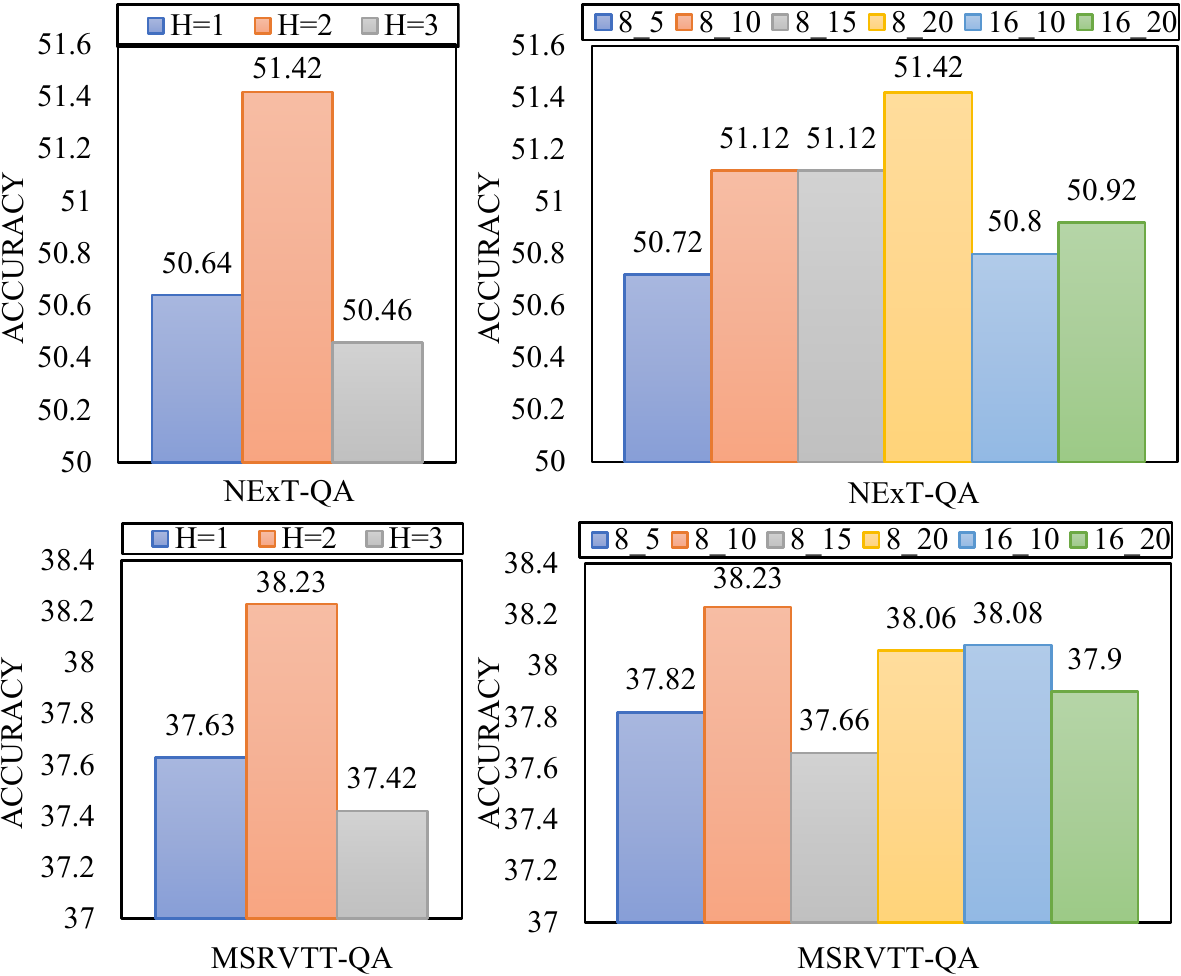}
 \caption{Investigation of graph attention layers H, sampled clips K and regions N on validation sets of NExT-QA and MSRVTT-QA. ($8\_5$ stands for $K=8$ and $N=5$.)}
 \label{fig:layer_cr}
 \vspace{-1.0em}
\end{figure}

\section{C. Model Analysis}

\subsection{Study of Hyper-Parameters}
As shown in Figure \ref{fig:layer_cr} (left), a two-layer graph attention $H=2$ in each QGA unit brings the best results for both datasets. We speculate that a one-layer graph is insufficient to learn the complex relations between different video elements, while a three-layer graph may over-smooth the visual components (as we stack 3 QGAs to achieve the whole hierarchical model) and hence hurt the performance. From Figure \ref{fig:layer_cr} (right), we can see that $K=8$ video clips are enough to bring good performances on the validation splits of both datasets. (Yet, we empirically find that our default setting of 16 clips yield the optimal results on NExT-QA test set.). However, NExT-QA demands relative more candidate regions ($N=20$) than MSRVTT-QA ($N=10$). Such difference is reasonable as NExT-QA's videos are relative longer with multiple objects interacted with each other.

\subsection{Analysis of Efficiency}
Our model is light-weight with three stacked 2-layer graphs, and graphs at the same level share parameters. Besides, modelling video as graph hierarchy benefits time efficiency compared with previous works that use RNNs. As videos are also sparsely sampled (4-16 clips, with 4 frames per clip) so running speed is fast (see Table \ref{tab:model}) once the features are ready. However, the overall speed (with feature extraction) could possibly be slower than those models that use only segment-level feature.

\subsection{Qualitative Analysis}

\begin{figure*}[t!]
 \centering
 \scalebox{0.8}{
     \subfloat[]{
     \includegraphics[width=1.0\textwidth]{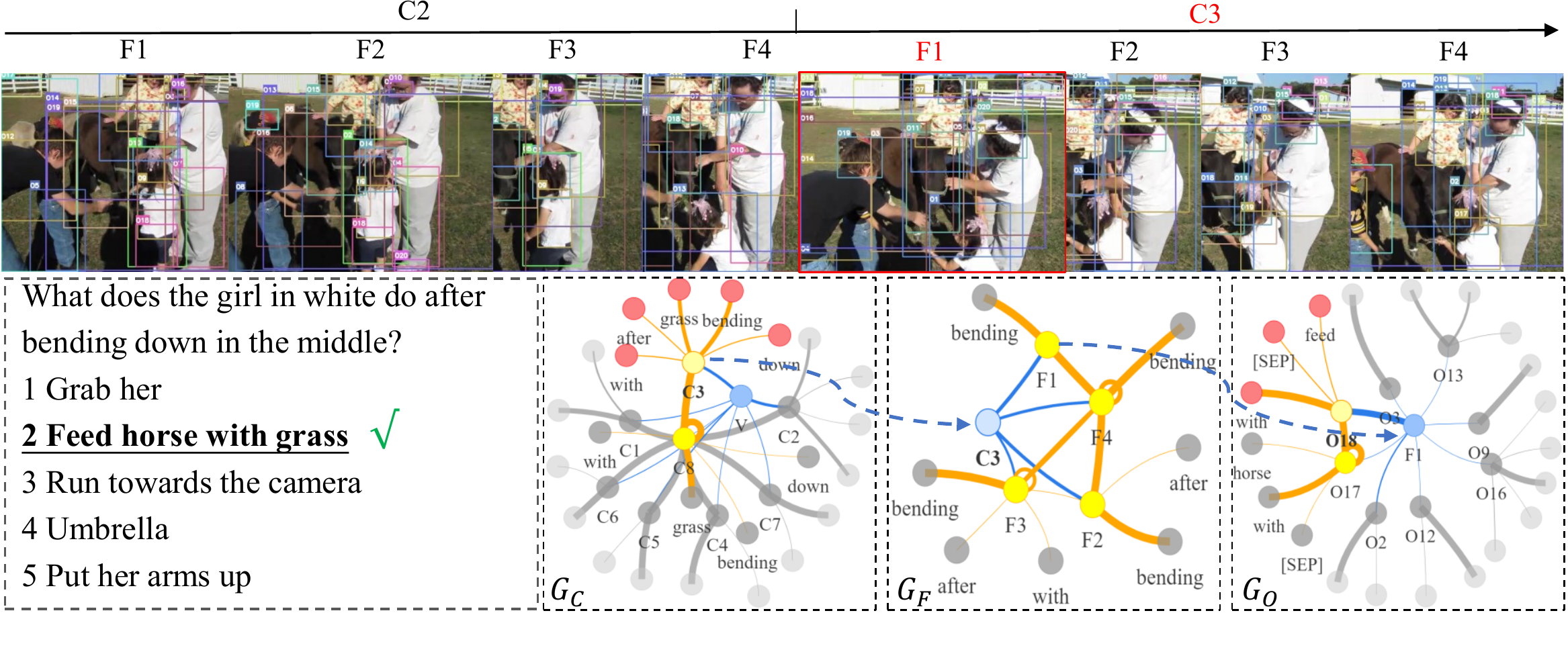}
     }
}
 \scalebox{0.8}{
     \subfloat[]{
     \includegraphics[width=1.0\textwidth]{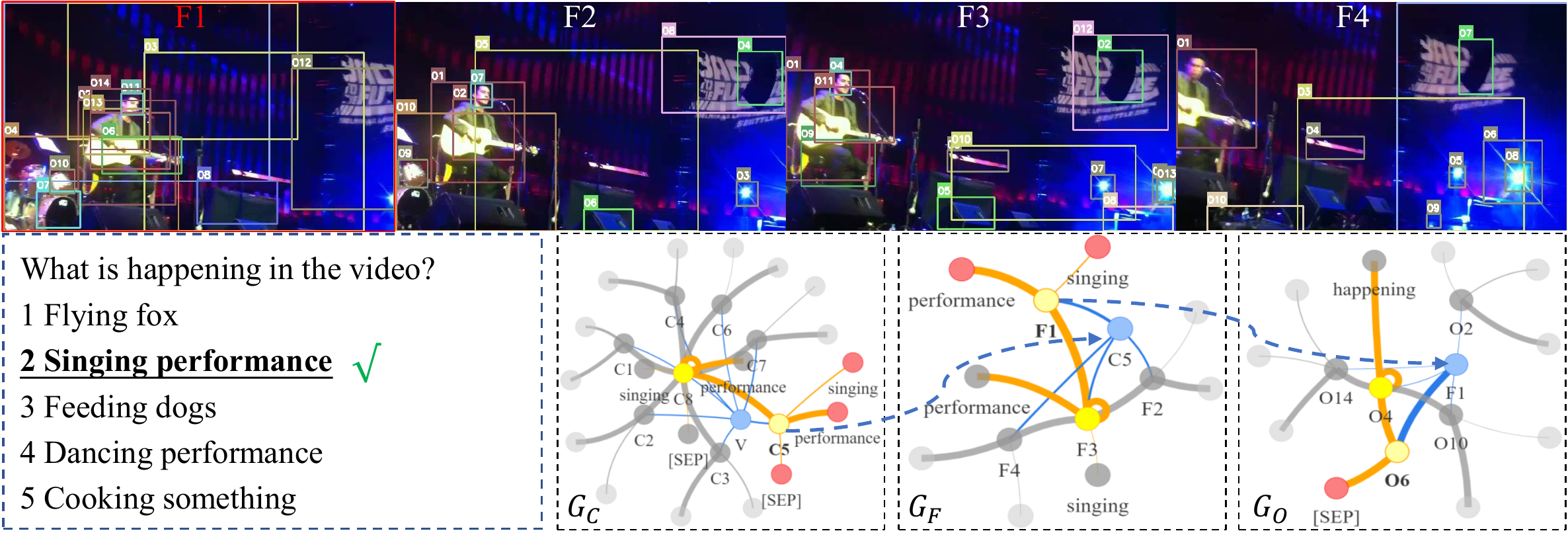}
    }
}
 \scalebox{0.8}{
    \subfloat[]{
     \includegraphics[width=1.0\textwidth]{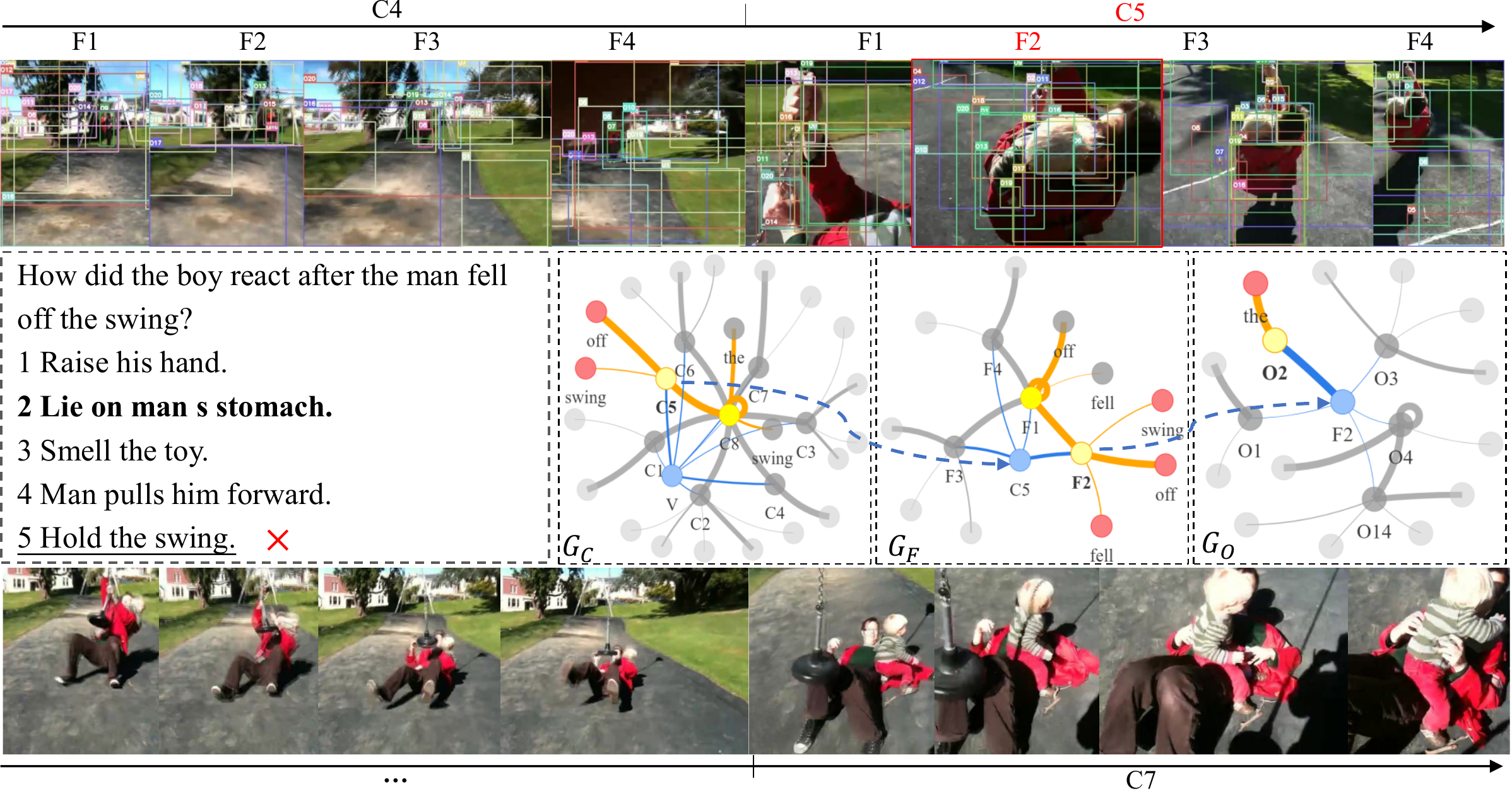}
    }
} 
 \caption{Visualization of question-answering results in NExT-QA \cite{xiao2021next}. The ground-truth answers and our predictions are highlighted in \textbf{bold} and \underline{underline} respectively.  \textcolor{blue}{Blue}: Self-attention weights $\beta$. \textcolor{orange}{Orange}: weights of adjacency matrix $A$ and query condition $\alpha$. (Please zoom in for better view of the object bounding boxes.)}
 \label{fig:quality-res}
 \vspace{-1.0em}
\end{figure*}

The example in Figure \ref{fig:quality-res}(a) shows that our model successfully predict the correct answer for the question that features both dynamic action (\eg, ``\texttt{bending down}'') and contact action (\eg, ``\texttt{feed horse}''). Meanwhile, the learned conditional graphs tell that our model can precisely align the query phrases of different contents with the video elements at the corresponding hierarchy level. For example, the expression ``\texttt{after bending down}'' which features dynamics is grounded at the top two levels, \ie, $G_C$ and $G_F$ that aim at capturing the interaction dynamics across frames and across clips respectively. In contrast, the expression ``\texttt{feed horse}'' which features contact action is grounded at the bottom level, \ie, $G_O$ that focuses on learning a static overview of the object interaction based on certain frames. Furthermore, by aligning the visual nodes in the graphs with the respective video contents, we can see that the model can spatio-temporally ground (localize) the question and answer in the video. Concretely, according to the two nodes $C_2$ and $C_3$ (in $G_C$) that response relatively stronger to the final video representation $V$, the model can temporally find the video clips that are relevant to the query. By tracing from $C_3$ through $F_1$ (in $G_F$) down to $O_{18}$ (in $G_O$), the model can further pinpoint the specific frame and region that are related to the answer ``\texttt{feed horse}''.

In Figure \ref{fig:quality-res}(b), we show another correct prediction case where the question gives no referring hint to the video contents but generally asks the global event in the video. To study why it can answer the question, again we trace down the learned graph hierarchy. We can see that the model can accurately finds the visual elements in the video to deduce the answer ``\texttt{singing performance}'', \eg, the singer ($O_{14}$, $O_2$), guitar ($O_6$) and other musical instruments ($O_4$, $O_{10}$) in frame $F_1$ of video clip $C_5$. The example demonstrates that the hierarchical architecture is able to abstract a set of low-level, local visual components into a high-level, global event even without explicit language cues.

Finally, we show a failing case in Figure \ref{fig:quality-res}(c). In the example, our model fails to obtain the correct answer ``\texttt{lie on man s stomach}'' but gets a negative one ``\texttt{hold the swing}''. By analyzing the wrong prediction, we find that the model 1) incorrectly aligns the video clips $C_4$ and $C_5$ with the referring expression ``\texttt{fell off swing}'' and 2) is also distracted by the negative answer ``\texttt{hold the swing}''. For the first problem, a further study of the video clips reveals that they are incomplete to cover the video contents corresponding to ``\texttt{fell off swing}'' which should be the contents between $C_6$ and $C_7$. We speculate that the process of ``fell off'' is likely transient and thus unfortunately gets missing during sampling in this case. Yet, our aforementioned study on video sampling suggests that 8 clips is enough for a overall good performance. For the second problem, we analyze it through the learned graph hierarchy. By tracing from the video clip $C_5$ (in $G_C$) through the video frame $F_2$ (in $G_F$) to the region $O_2$ (in $G_O$), we find that the region content $O_2$ does correspond to the distractor answer ``\texttt{hold the swing}''. Besides, the contents of video clip $C_5$ are visually similar to ``fell off''. Consequently, the two factors jointly lead our model to the wrong answer.

Overall, the above analyses demonstrate the effectiveness of our hierarchical conditional graph model in 1) reasoning and inferring video elements of different granularity, and 2) fine-grained matching between visual and textual contents at multi-granularity levels. Also, the model is of enhanced explanability either in interpreting the correct predictions, or in diagnosing the source of the wrong predictions.

\clearpage
\pagebreak
\end{document}